# DScheLLM: Enabling Dynamic Scheduling through a Fine-Tuned Dual-System Large language Model

Lixiang Zhang[a,b], Chenggong Zhao[b], Qing Gao[a,b,*], Xiaoke Zhao[b], Gengyi Bai[b], Jinhu Lv[a]

[a] *School of Automation Science and Electrical Engineering, Beihang University, Beijing, P.R. China*
[b] *Hangzhou Innovation Institute, Beihang University, Hangzhou, P.R. China*

**Abstract**

Production scheduling is highly susceptible to dynamic disruptions, such as variations in processing times, machine availability, and unexpected task insertions. Conventional approaches typically rely on event-specific models and explicit analytical formulations, which limits their adaptability and generalization across previously unseen disturbances. To overcome these limitations, this paper proposes DScheLLM, a dynamic scheduling approach that leverages fine-tuned large language models within a dual-system (fast–slow) reasoning architecture to address disturbances of different scales. A unified large language model-based framework is constructed to handle dynamic events, where training datasets for both fast and slow reasoning modes are generated using exact schedules obtained from an operations research solver. The Huawei OpenPangu Embedded-7B model is subsequently fine-tuned under the hybrid reasoning paradigms using LoRA. Experimental evaluations on standard job shop scheduling benchmarks demonstrate that the fast-thinking mode can efficiently generate high-quality schedules and the slow-thinking mode can produce solver-compatible and well-formatted decision inputs. To the best of our knowledge, this work represents one of the earliest studies applying large language models to job shop scheduling in dynamic environments, highlighting their considerable potential for intelligent and adaptive scheduling optimization.

## 1. Introduction

Production scheduling, particularly the job shop scheduling problem (JSP), has long been recognized as a fundamental and challenging topic in manufacturing systems research. As an NP-hard combinatorial optimization problem, JSP involves determining feasible processing sequences for multiple jobs competing for limited machine resources under precedence constraints. Over the past decades, extensive research efforts, including exact optimization methods, heuristics, metaheuristics, and learning-based approaches, have led to significant progress. However, most existing studies focus on designing tailored algorithms for specific objectives or constraint settings, resulting in a fragmented solution landscape that limits transferability and scalability in practical industrial applications.

In real-world shop-floor environments, dynamic events such as job insertions or cancellations, machine breakdowns, and processing-time variations frequently occur and can invalidate original schedules or severely degrade system performance, thereby necessitating timely and reliable rescheduling. To address these challenges, numerous dynamic scheduling approaches have been proposed, including reactive heuristics, robust optimization techniques, and predictive-reactive hybrid frameworks. While some methods emphasize rapid responsiveness to minimize operational disruptions, others focus on maintaining global optimality under uncertainty. Nevertheless, these approaches often exhibit limited adaptability when facing heterogeneous, compound, or evolving disruptions, and typically require algorithm redesign or parameter retuning for new scenarios. First, most approaches lack a unified mechanism capable of addressing diverse dynamic events, including previously unseen disruptions, without relying on event-specific models or heuristics. This limitation leads to a proliferation of specialized solutions with restricted adaptability in environments where novel disturbances frequently arise. Second, existing scheduling models generally provide limited interpretability. Mathematical optimization and combinatorial solvers often demand substantial domain expertise for configuration and analysis, and the underlying rationale for rescheduling decisions is difficult to express in an intuitive or human-readable manner, hindering effective human-machine collaboration.

To overcome these challenges, this paper proposes DScheLLM, a dynamic scheduling framework that leverages fine-tuned large language models (LLMs) with a dual-system (fast-slow) reasoning mechanism to address disturbances of varying complexity and scale. The proposed framework makes three primary contributions.

1) It establishes a general and extensible scheduling framework capable of handling single-event, multi-event, and previously unseen disruptions, thereby enhancing robustness and generalization in highly dynamic environments.
2) By fine-tuning an LLM with explicit chain-of-thought reasoning, the framework enables direct inference of rescheduling decisions from original schedules and disturbance descriptions expressed in natural language, substantially improving interpretability for human operators.
3) It introduces a dual-system reasoning mechanism, comprising a fast-thinking mode for rapid and interpretable schedule adjustments suitable for real-time response, and a slow-thinking mode that performs stepwise reasoning to address complex multi-event disruptions, generating solver-compatible scheduling representations for integration with mathematical optimization solvers.

## 2. Related work

### 2.1. Solving Methods on Dynamic JSP

This study focuses on the dynamic variant of the problem, where scheduling must accommodate not only the static constraints but also dynamic events such as processing time changes, due date variations, job arrivals, machine breakdowns, and preventive maintenance. These disruptions give rise to the dynamic job shop scheduling problem (DJSP), which requires schedules to be highly adaptive and responsive in real time. To address the challenges posed by dynamic shop-floor environments, current research tends to rely on heuristic rule–based strategies and RL-driven real-time scheduling methods [1], [2].

In the domain of heuristic-rule learning, Eguchi et al. [3] proposed a neural-network-based priority rule learning method that demonstrated strong scheduling performance across scenarios with varying machine utilizations and due-date tightness. Nguyen and Johnston [4] developed a local-search-based automatic planning algorithm to evolve heuristic rules, achieving better performance than tree-based genetic programming and gene expression programming. Although heuristic-rule learning approaches achieve promising results in certain settings, they often suffer from high design complexity and poor scalability in practical deployment. RL-based approaches have gained prominence as a means to achieve more efficient real-time scheduling, especially in balancing optimization quality and responsiveness. Wang [5], for example, introduced a weighted Q-learning–based adaptive scheduling strategy that extracted state features through clustering and updated Q-function weights iteratively to accelerate learning and improve accuracy. Aydin and Oztemel [6] proposed an RL-driven dispatching rule selection mechanism for DJSPs, demonstrating its effectiveness in handling dynamic job arrivals and processing-time fluctuations.

Because traditional Q-learning struggles with large-scale continuous state spaces, researchers have increasingly incorporated deep learning into RL, giving rise to deep reinforcement learning (DRL) approaches. Wang et al. [7] developed a DRL-based real-time scheduling method that improved both stability and accuracy in policy learning. Zhao et al. [8] proposed a deep Q network-based scheduling framework with manually engineered state features and dispatching-rule outputs. Their method demonstrated stronger adaptability than conventional Q-learning and heuristic rules. Building on this direction, Zhou et al. [9] introduced an online DRL scheduling method capable of continuously adapting scheduling policies in response to uncertain events such as dynamic order arrivals. Researchers have also explored policy-based and hybrid value-policy DRL methods. Liu et al. [10], for instance, proposed an actor–critic scheduling framework using deep deterministic policy gradients (DDPG) to train multiple agents associated with different dispatching rules in parallel, showing robust performance in both static and dynamic settings. Wang and Liao [11] combined discrete-event simulation with proximal policy optimization (PPO) to develop a real-time scheduling method that outperformed rule-based and heuristic methods in both optimization quality and responsiveness when solving DJSPs.

To further enhance the adaptability of scheduling policies, Liu et al. [12] introduced a graph reinforcement learning method in which a hybrid graph transformer network (GTN) was used to extract state embeddings reflective of dynamic events. Combined with replay techniques to improve sampling efficiency in PPO, this method demonstrated competitive performance on dynamic scheduling tasks. Additionally, Liu et al. [13] developed a centralized-training, decentralized-execution multi-agent RL framework that showed superior performance when solving variable-scale DJSPs.

*2.2. LLMs in production*

Owing to LLMs' strong representation learning and reasoning capabilities, LLMs have been explored for a variety of manufacturing-related tasks, including process planning, production decision support, and simulation modelling [14]. Existing studies can be broadly categorized into two groups: LLMs as auxiliary components for existing methods, and LLMs as direct solvers or generators of solutions.

As for assistant components, Hong and Li [15] proposed the LLM4A3C framework, which integrates an LLM with the A3C algorithm to address the flexible job shop scheduling problem by semantically enhancing state representations and generating potential-based reward functions for adaptive optimization. Wang et al. [16] developed the multi-agent scheduling chain, leveraging LLM-driven scheduling agents and dialogue-based knowledge retention to enable dynamic and real-time rescheduling. Xu et al. [17] employed LLMs to construct machining process knowledge graphs, supporting intelligent process planning through high-quality data annotation and knowledge extraction.

In contrast, another line of research explores the direct use of LLMs to generate solutions, plans, or executable models with minimal reliance on conventional optimization pipelines. Abgaryan et al. [18] addressed the JSP by fine-tuning LLMs on the large-scale Starjob dataset, enabling end-to-end scheduling directly from problem instances. Tsushima et al. [19] developed an LLM-based robot work planning system that directly generates executable robot tasks through natural language interaction. Hu et al. [20] proposed a prompt-engineering-based approach in which LLMs directly generate effective shop floor layouts via an iterative reasoning and evaluation process. Elbasheer et al. [21] introduced an LLM-enabled framework that directly transforms natural language conversations into executable simulation models, bypassing manual model specification. Finally, Ni et al. [22] presented an LLM-based adaptive process management approach that directly converts informal user descriptions into structured and validated manufacturing workflows.

*2.3. Literature Review Summary and Research Motivation*

In summary, prior research on DJSP has produced a wide range of effective methods. However, most existing approaches exhibit limited generalization ability, poor scenario transfer, and strong dependence on problem-specific modelling. At the same time, recent studies on LLMs in manufacturing demonstrate their potential to encode domain knowledge and support complex decision-making processes, yet their use in scheduling has largely been confined to auxiliary or prompt-based roles. The integration of LLM fine-tuning with dynamic scheduling remains fragmented and incomplete.

These observations reveal a clear research gap: the lack of a general-purpose, fine-tuned large language model capable of directly addressing dynamic scheduling problems across diverse and evolving environments. Motivated by this gap, this study aims to develop a fine-tuned LLM that can internalize scheduling knowledge and adapt to dynamic job shop scenarios without task-specific modelling. By bridging the gap between dynamic scheduling theory and foundation model learning, this work seeks to advance the development of scalable and transferable scheduling solutions for complex manufacturing systems.

## 3. Methods

### 3.1. Problem Description

The JSP is a classical combinatorial optimization problem that involves scheduling a set of jobs on a set of machines under predefined technological constraints. Let $N = \{1, 2, \ldots, n\}$ denote the set of jobs and $M = \{1, 2, \ldots, m\}$ denote the set of machines, where $n$ and $m$ are the numbers of jobs and machines, respectively. Each job $i \in N$ consists of $n_i$ operations that must be processed in a given order. The operation $O_{ik}$ represents the $k$-th operation of job $i$, which must be processed on a specified machine for a known processing time $p_{ik}$. Let $s_{ik}$ denote the starting time of operation $O_{ik}$. The precedence constraints within each job are defined by a set of ordered operation pairs, ensuring that each operation cannot start until its predecessor has been completed. In addition, each machine can process at most one operation at a time, and all operations assigned to the same machine form a machine-specific operation set. Typically, both the starting times and processing times are assumed to be integer-valued.

The objective is to minimize the makespan in (1), defined as the maximum completion time among all jobs, where $C_i$ represents the completion time of job $i$.

$$\min C_{\max} = \min \left( \max_{i \in N} C_i \right) \tag{1}$$

The scheduling solution is subject to two types of constraints: sequence constraints and resource constraints.

Sequence constraints require that all operations of each job be processed in a predetermined order, expressed as (2).

$$s_{i,k+1} \geq s_{i,k} + p_{i,k}, \forall i \in N, k = 1, 2, \cdots, n_i - 1 \tag{2}$$

Resource constraints ensure that each machine can process at most one operation at any given time. When two operations $O_{ik}$ and $O_{jl}$ are assigned to the same machine, the sequence variable is defined as (3). The resource constraint is formulated as (4).

$$x_{ik,jl} = \begin{cases} 1, O_{ik} \text{ is processed before } O_{jl} \\ 0, \text{otherwise} \end{cases} \tag{3}$$

$$\begin{cases} s_{i,k} \geq s_{j,l} + p_{j,l}, \text{if } x_{ik,jl} = 0 \\ s_{j,l} \geq s_{i,k} + p_{i,k}, \text{otherwise} \end{cases} \tag{4}$$

In this study, five types of dynamic events are considered, including job insertion, job cancellation, processing time variation, machine assignment change, and machine maintenance. Among these, only machine maintenance directly affects the structure of the JSP optimization model. The maintenance of machine $m$ introduces an unavailable time interval in (5), during which the machine cannot process any operation.

$$UT_m = [mst_m, met_m] \qquad (5)$$

Accordingly, an additional resource constraint is introduced in (6). For any operation $O_{ik}$ assigned to machine $m$, its processing interval must not overlap with the maintenance period.

$$\left[ s_{i,k}, s_{i,k} + p_{i,k} \right] \cap UT_m = \varnothing, if\ M_{i,k} = m \qquad (6)$$

### 3.2. Fine-Tuning Framework

To address the limitations of traditional scheduling methods in handling complex and highly dynamic events, this paper proposes an LLM-based optimization framework for DJSP, as shown in Fig. 1. The proposed framework consists of three main components: dataset generation, model fine-tuning, and model deployment.

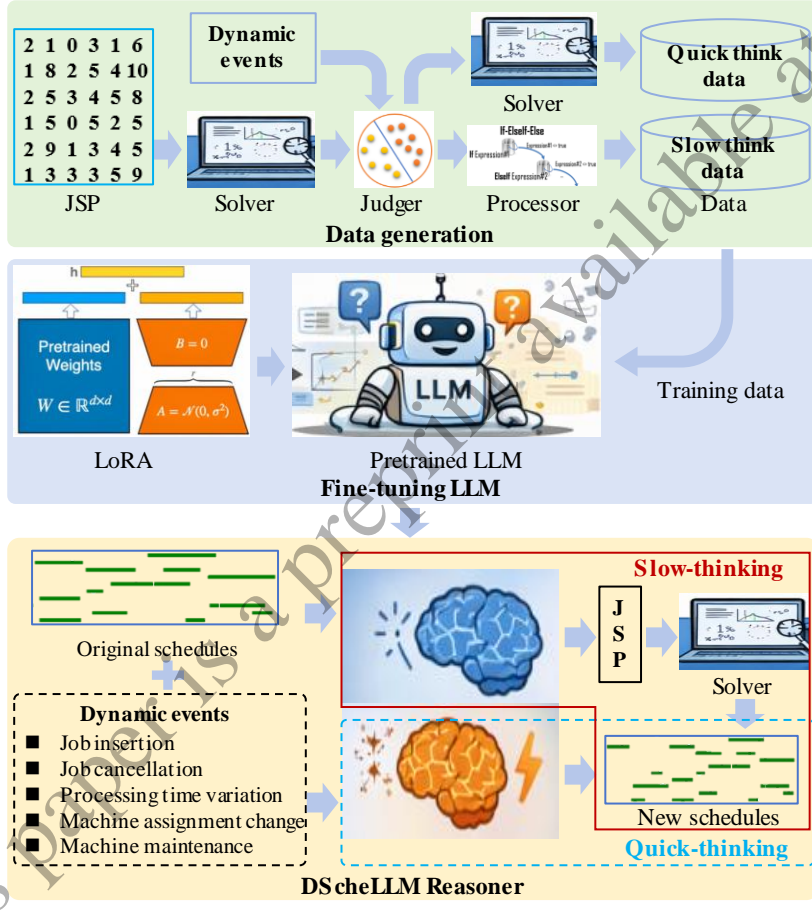

Fig. 1. The framework of training the DScheLLM

In the dataset generation stage, an operations research solver-driven approach is adopted to generate the original schedules. Then, random dynamic events are deliberately injected into the original scheduling environment. A specially designed "***Judger***" module is then introduced to evaluate the degree of disruption caused by dynamic events to the existing schedule. When the disturbance is minor, the problem is classified as a simple task. In this case, the solver is directly invoked to generate the updated schedules, and the resulting input-output pairs are used to construct a fast-thinking dataset. Conversely, when the disturbance is significant, the problem is categorized as a complex task. For such cases, a logic-based "***Processor***" is manually designed to infer a standardized and structured description of the JSP, which can be directly consumed by the solver. The specific data format is detailed in Section 3.3. This dual-path data generation strategy enables the construction of high-quality datasets that explicitly encode different levels of reasoning complexity.

To reduce computational overhead while effectively leveraging the reasoning capabilities of pretrained LLMs, this work adopts the low-rank adaptation (LoRA) technique to fine-tune the Huawei OpenPangu Embedded-7B model. The fine-tuning process is conducted on a mixed dataset consisting of both fast-thinking and slow-thinking samples, with 10,000 instances in each subset. Through this targeted adaptation strategy, a domain-specific model, referred to as DScheLLM, is obtained, enabling adaptive reasoning in dynamic job shop scheduling scenarios.

During deployment, DScheLLM automatically evaluates the complexity of incoming scheduling instances and selects an appropriate reasoning mode by activating the corresponding parameter subspaces. When dynamic events induce only minor perturbations to the original schedule, the model operates in a fast-thinking mode and directly generates revised scheduling solutions. Conversely, under significant disruptions, DScheLLM switches to a slow-thinking mode, producing a standardized and structured JSP representation. This representation is subsequently passed to a pre-deployed operations research solver, implemented as a constraint programming model in Google OR-Tools, to compute optimal or near-optimal schedules.

By combining adaptive LLM-based reasoning with traditional optimization solvers, the fine-tuned DScheLLM is able to respond rapidly and accurately to a wide range of dynamic disturbances in job shop scheduling environments. This integrated framework substantially reduces reliance on expert knowledge and manual intervention, thereby lowering the barrier to practical deployment and enabling efficient and scalable application in real-world manufacturing systems.

*3.3. Dataset Generation*

To enable the LLM to effectively respond to dynamic events and efficiently solve JSPs, the training dataset must incorporate well-defined optimization objectives and precise constraint specifications. Each input instance consists of two components as shown in Fig. 2: (i) an instruction prompt and (ii) a problem description that includes the current schedule and the corresponding dynamic events.

The instruction explicitly specifies the expected outputs of the LLM, as illustrated below:

*"You are given a current schedule for a Job Shop Scheduling Problem (JSSP), as well as one or more dynamic events. Based on the impact of the dynamic event, you are required to: 1. Generate updated scheduling tasks. 2. Generate machine unavailability intervals. 3. If the dynamic event has a minor impact on the current schedules, producing locally adjusted schedules that adapts the original schedule to these changes while minimizing disruption to unaffected operations."*

When the LLM is required to directly output a locally adjusted schedule, a set of constraint descriptions is provided to strictly regulate the solution space. These constraints are described as follows.

*"The locally adjusted schedules must strictly satisfy the following constraints: 1. Each operation must be processed exactly once. 2. All operations of each job must be processed in the given order. 3. Each operation must be processed on its designated machine. 4. Each machine can process at most one operation at any given time. 5. Preemption is not allowed: once an operation starts, it must be processed continuously until completion. 6. No operation may be scheduled during machine unavailability intervals. 7. The adjustment should be local, meaning that only operations affected by the dynamic events should be rescheduled whenever possible, while preserving the original schedule for unaffected operations."*

The instruction component and the overall format of the problem description are identical for both fast-thinking and slow-thinking modes. The primary difference lies in the complexity of the dynamic event specification. In the fast-thinking setting, each instance involves only a single type of dynamic disturbance, and a special identifier "**/no_think**" is appended to the end of the problem description. In contrast, the slow-thinking setting may include two or more types of dynamic temporal disturbances, thereby increasing problem complexity.

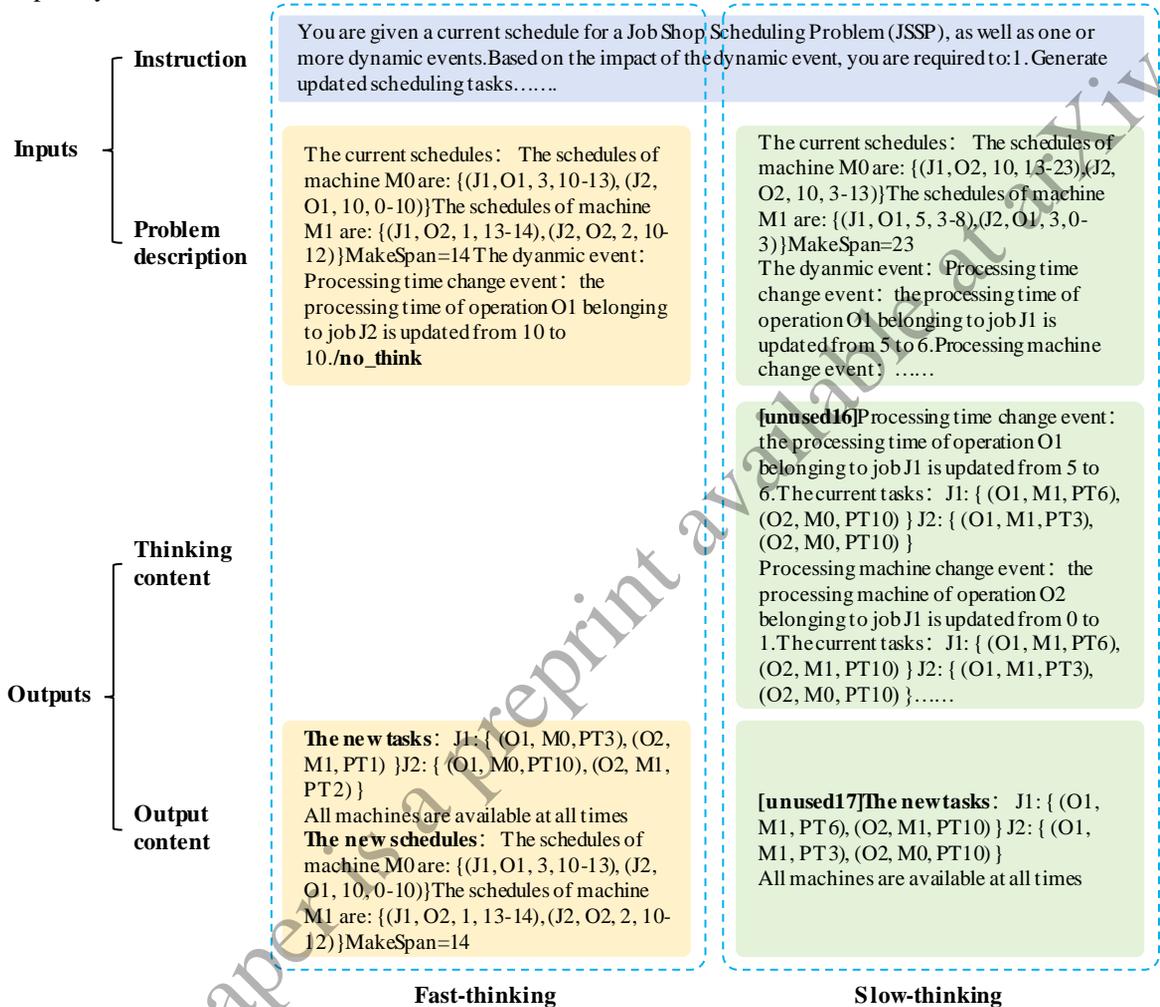

Fig. 2. The format of datasets for fine-tuning

The output design further distinguishes the two reasoning modes. In the fast-thinking mode, the model directly outputs the updated, standardized JSP instance together with the corresponding revised schedule. In the slow-thinking mode, the output contains both an explicit reasoning process and the final standardized JSP, denoted by the special tokens "**[unused16]**" and "**[unused17]**", respectively. The reasoning content is manually structured to sequentially address five categories of dynamic events. After processing each event, an updated standardized JSP is generated. This step-by-step representation enables the LLM to learn the reasoning patterns that link dynamic disruptions to scheduling transformations, thereby enhancing its ability to handle complex dynamic scheduling scenarios.

Based on the above design, this study constructs 10,000 instances for both fast-thinking and slow-thinking settings, with detailed statistics summarized in Table 1. The numbers of jobs and machines are independently sampled from discrete uniform distributions. In the fast-thinking setting, a single dynamic event is randomly selected from five predefined event types. In contrast, the slow-thinking setting includes multiple dynamic events, with at least two events occurring in each instance. Specifically, the numbers of job cancellation events (*NJC*), processing time change events (*NTC*), processing machine change events (*NMC*), job insertion events (*NJI*), and machine maintenance events (*NMM*) are all drawn from discrete uniform distributions.

Table 1. The setting of instances for fine-tuning

| Mode | Fast-thinking | Slow-thinking |
| --- | --- | --- |
| The number of jobs | $U\{2, 6\}$ | $U\{2, 6\}$ |
| The number of machines | $U\{2, 6\}$ | $U\{2, 6\}$ |
| The number of dynamic events | 1 | $NJC \sim U\{0, 1\}$ <br> $NTC \sim U\{1, 2\}$ <br> $NMC \sim U\{1, 2\}$ <br> $NJI \sim U\{0, 2\}$ <br> $NMM \sim U\{0, 2\}$ |
| The number of instances | 10,000 | 10,000 |

*3.4. Parameter-Efficient Fine-Tuning with LoRA*

To substantially reduce the computational and memory overhead associated with supervised fine-tuning while preserving the performance of pretrained LLMs, this work adopts the parameter-efficient fine-tuning method known as LoRA. LoRA injects a small number of low-rank trainable parameters into the key linear projection layers of a pretrained Transformer model, while keeping all original pretrained weights frozen. This design enables efficient adaptation to downstream tasks with only a marginal increase in computational cost.

Let the parameter set of the pretrained model be denoted by (7), where $W_l$ represents the linear projection weights of the *l*-th layer.

$$\Theta_0 = \{W_1, W_2, \ldots, W_L\} \tag{7}$$

Unlike conventional full-parameter fine-tuning, which directly updates all parameters in $\Theta_0$, LoRA introduces a low-rank parameter increment $\Delta W_l$ such that the adapted model parameters can be expressed as (8). By constraining parameter updates to a low-rank subspace, LoRA significantly reduces the number of trainable parameters while retaining the expressive capacity of the original model.

$$\Theta = \Theta_0 + \Delta\Theta \tag{8}$$

The proposed method is built upon a pretrained Transformer-based language model. For an arbitrary linear transformation within the model, the original computation can be expressed as (9), where *x* denotes the input features and $W_0$ represents the frozen pretrained weight matrix.

$$h = W_0 x \tag{9}$$

Under the LoRA framework, this linear transformation is augmented with a low-rank update and reformulated as (10), where *A* and *B* are trainable low-rank matrices, and *α* is a fixed scaling factor that controls the contribution of the low-rank update to the output. Across the entire model, only the low-rank parameters are optimized during training, while all pretrained weights remain unchanged. This design enables parameter-efficient adaptation without modifying the original model architecture.

$$h = W_0 x + \alpha A B x \tag{10}$$

The model is trained using a supervised fine-tuning paradigm. Each training sample consists of an input prompt concatenated with its corresponding generated response, forming a complete input sequence to the model. To avoid ineffective optimization over the prompt tokens, a masking strategy is applied during loss computation such that the cross-entropy loss is evaluated only on the generated response tokens. Let the sequence length be $T$, the training loss is defined as (11), Where $y_t$ is the ground-truth token at position $t$, and $m_t$ is a binary mask variable, with $m_t = 0$ for prompt tokens and $m_t = 1$ for response tokens.

$$\mathcal{L} = -\sum_{t=1}^{T} m_t \cdot \log P(y_t | x) \tag{11}$$

During training, only the low-rank parameters introduced by LoRA are optimized by minimizing the above loss function, while all pretrained model parameters remain frozen. This training strategy guides the model to generate high-quality responses conditioned on the given prompts in a parameter-efficient manner.

*3.5. Training*

Supervised fine-tuning is conducted on Huawei OpenPangu-7B-v1.1 [23], a pretrained language model with 7 billion parameters developed by Huawei. The LoRA rank is set to $r = 16$, with a scaling factor of $\alpha = 32$ and a dropout rate of 0.05.

Input sequences are constructed using a unified prompt template, ensuring consistent formatting across all training samples. During loss computation, prompt tokens are masked so that the cross-entropy loss is evaluated exclusively on the generated response tokens, thereby focusing optimization on task-relevant outputs. Training is performed for three epochs on eight Ascend NPUs using the Adam optimizer with an initial learning rate of $1 \times 10^{-4}$. FP16 mixed-precision training is employed to improve computational efficiency, and gradient norm clipping is applied to maintain numerical stability during optimization. For reproducibility and robust model selection, both the checkpoint achieving the best validation performance and the final checkpoint are retained. The training loss curve during fine-tuning is depicted in Fig. 3. The loss demonstrates stable optimization and the absence of training instabilities during fine-tuning.

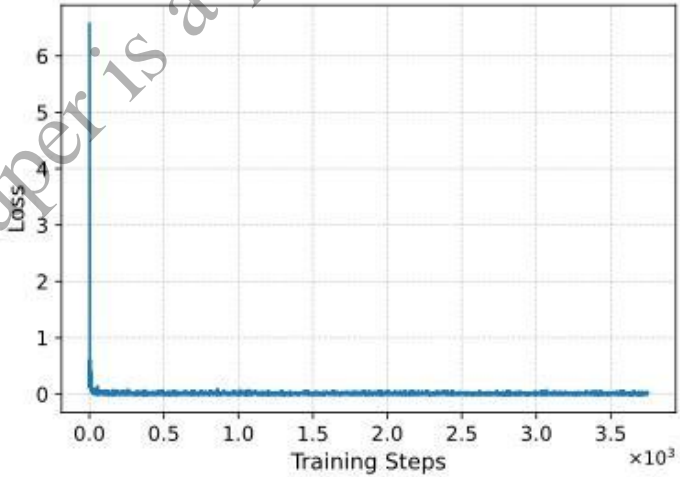

Fig. 3. Training loss curve during fine-tuning

## 4. Experiments

### 4.1. Instances

To rigorously evaluate the fine-tuned LLM proposed in this study, experiments are conducted on the classical FT06 JSP benchmark, originally proposed by Fisher and Thompson. The instruction format follows the specification presented in the previous sections. The datasets are set as shown in Fig. 4. First, the operations research-based solver developed in this work is used to compute the optimal schedule for the original FT06 instance, which is then converted into a natural-language description of the scheduling solution.

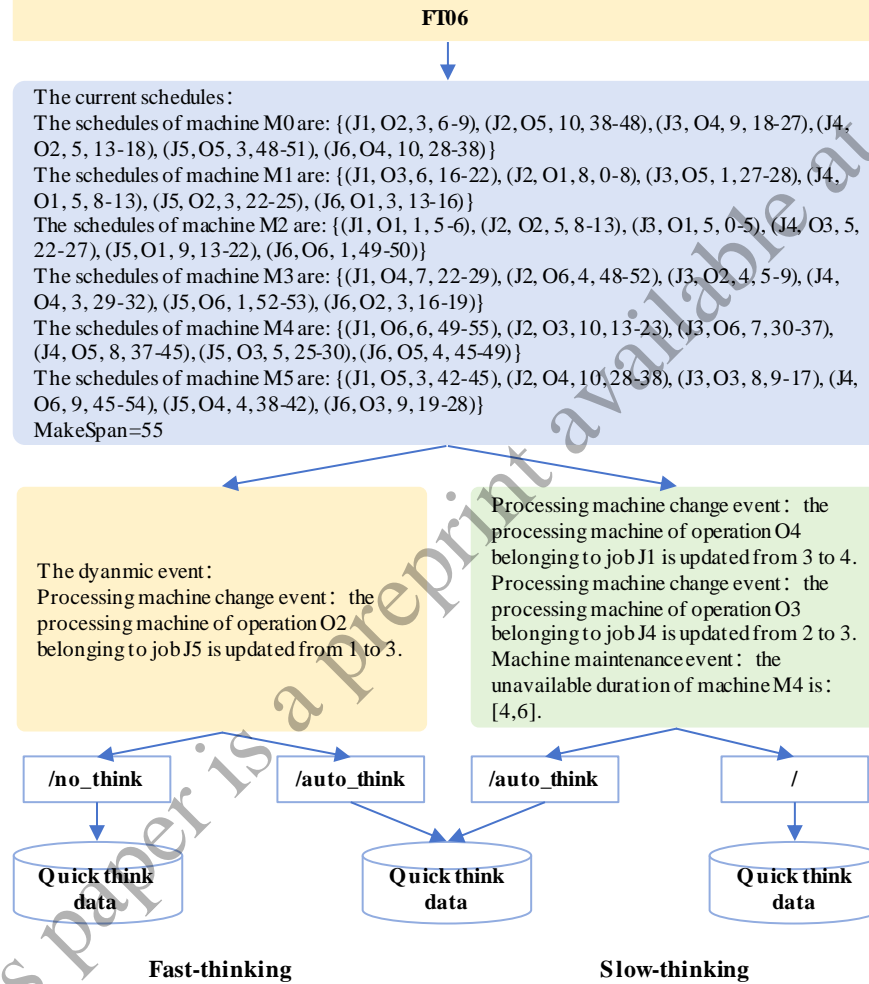

Fig. 4. The format of datasets for testing

For the fast-thinking evaluation, a single dynamic event is randomly sampled from a discrete uniform distribution. A fast-thinking instance is generated by appending the identifier "**no_think**" to the input. In addition, an automatic-thinking variant is constructed by appending the identifier "**auto_think**" enabling evaluation of the model's autonomous reasoning-mode selection capability. For the slow-thinking evaluation, the number of dynamic events is first sampled from a discrete uniform distribution over five event types, as shown in Table 1, and the specific dynamic events are then generated sequentially. An automatic-thinking instance for slow-thinking problems is created by appending the identifier "**auto_think**" to the input. When no identifier is appended, the model defaults to slow-thinking mode.

For each experimental scenario, 30 instances are generated for systematic evaluation. Together, these settings provide a comprehensive assessment of the model's adaptive reasoning capabilities across both fast- and slow-thinking scenarios.

*4.2. Evaluation on Automatic Modes*

As shown in Table 2, the fine-tuned LLM was evaluated under the automatic-thinking mode. For fast-thinking tasks, the automatic mode achieved 100% accuracy in selecting the appropriate reasoning strategy. In contrast, for slow-thinking tasks, the accuracy decreased to 33.33%, suggesting that the pretrained model's built-in fast-slow reasoning discriminator may be insufficient for accurately assessing problem complexity in this industrial scheduling scenario. These results indicate that, although the automatic mode is reliable for relatively simple and fast-response tasks, it struggles to identify situations that require deeper reasoning. Consequently, further research is needed to improve the automatic reasoning-mode selection mechanism, enabling more accurate adaptation to complex combinatorial and multi-event scheduling problems.

Table 2. Accuracy of automatic reasoning modes

| Accuracy | Fast-thinking | Slow-thinking |
| --- | --- | --- |
| Fast-thinking datasets | 100% | / |
| Slow-thinking datasets | 66.67% | 33.33% |

*4.3. Evaluation on Fast-thinking*

As illustrated in Fig. 5, the fine-tuned LLM demonstrates strong performance on the fast-thinking tasks. Specifically, 46.67% of the generated solutions are optimal, while 73.33% are feasible, indicating that the model produces valid scheduling solutions in most test scenarios under limited inference time. Further analysis shows that, across the five types of dynamic events considered, the model exhibits robust situational awareness and rapid decision-making in the fast-thinking mode. Upon the occurrence of dynamic events, it can promptly adapt to changes in the system state and generate updated scheduling plans.

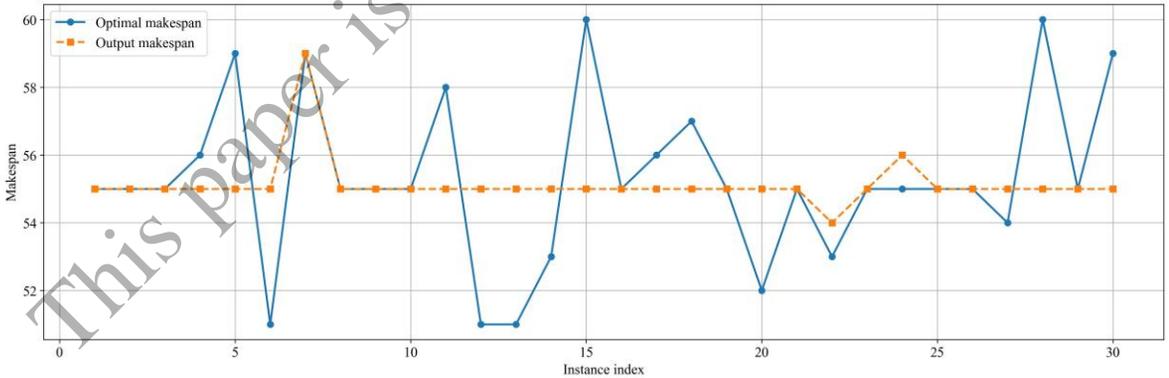

Fig. 5. Results of fast-thinking

Although the fast-thinking mode prioritizes inference efficiency over global optimality, nearly half of the generated solutions still achieve optimal performance. This observation suggests that the proposed fine-tuning strategy preserves the model's understanding of the underlying problem structure and scheduling constraints while enabling rapid response. Overall, these results demonstrate the feasibility and effectiveness of the fast-thinking mode for dynamic scheduling scenarios and establish a basis for further performance enhancement through integration with the slow-thinking mode.

*4.4. Evaluation on Slow-thinking*

As shown in Fig. 6, the fine-tuned large language model exhibits strong performance in the slow-thinking mode. The model is able to handle diverse combinations of dynamic events, and all generated outputs are compatible with the operations research solver for subsequent optimization. This result indicates that the model satisfies the reasoning requirements of complex dynamic scheduling tasks while ensuring solution feasibility and structural consistency across different event combinations.

=== Input ===
The current schedules： The schedules of machine M0 are: {(J1, O2, 3, 6-9),(J2, O5, 10, 38-48),(J3, O4, 9, 18-27),(J4, O2, 5, 13-18),(J5, O5, 3, 48-51),(J6, O4, 10, 28-38)}The schedules of machine M1 are: {(J1, O3, 6, 16-22),(J2, O1, 8, 0-8),(J3, O5, 1, 27-28),(J4, O1, 5, 8-13),(J5, O2, 3, 22-25),(J6, O1, 3, 13-16)}The schedules of machine M2 are: {(J1, O1, 1, 5-6),(J2, O2, 5, 8-13),(J3, O1, 5, 0-5),(J4, O3, 5, 22-27),(J5, O1, 9, 13-22),(J6, O6, 1, 49-50)}The schedules of machine M3 are: {(J1, O4, 7, 22-29),(J2, O6, 4, 48-52),(J3, O2, 4, 5-9),(J4, O4, 3, 29-32),(J5, O6, 1, 52-53),(J6, O2, 3, 16-19)}The schedules of machine M4 are: {(J1, O6, 6, 49-55),(J2, O3, 10, 13-23),(J3, O6, 7, 30-37),(J4, O5, 8, 37-45),(J5, O3, 5, 25-30),(J6, O5, 4, 45-49)}The schedules of machine M5 are: {(J1, O5, 3, 42-45),(J2, O4, 10, 28-38),(J3, O3, 8, 9-17),(J4, O6, 9, 45-54),(J5, O4, 4, 38-42),(J6, O3, 9, 19-28)}MakeSpan=55 The dyanmic event： Processing time change event： the processing time of operation O1 belonging to job J3 is updated from 5 to 10.Processing time change event： the processing time of operation O3 belonging to job J4 is updated from 5 to 9.Processing machine change event： the processing machine of operation O1 belonging to job J5 is updated from 2 to 0.Processing machine change event： the processing machine of operation O6 belonging to job J6 is updated from 2 to 0.Machine maintenance event： the unavailable duration of machine M1 is： [8,11].

=== Model Output ===
[unused16]Processing time change event： the processing time of operation O1 belonging to job J3 is updated from 5 to 10.The current tasks： J1: { (O1, M2, PT1), (O2, M0, PT3), (O3, M1, PT6), (O4, M3, PT7), (O5, M5, PT3), (O6, M4, PT6) } J2: { (O1, M1, PT8), (O2, M2, PT5), (O3, M4, PT10), (O4, M5, PT10), (O5, M0, PT10), (O6, M3, PT4) } J3: { (O1, M2, PT10), (O2, M3, PT4), (O3, M5, PT8), (O4, M0, PT9), (O5, M1, PT1), (O6, M4, PT7) } J4: { (O1, M1, PT5), (O2, M0, PT5), (O3, M2, PT5), (O4, M3, PT3), (O5, M4, PT8), (O6, M5, PT9) } J5: { (O1, M2, PT9), (O2, M1, PT 3), (O3, M4, PT5), (O4, M5, PT4), (O5, M0, PT3), (O6, M3, PT1) } J6: { (O1, M1, PT3), (O2, M3, PT3), (O3, M5, PT9), (O4, M0, PT10), (O5, M4, PT4), (O6, M2, PT1) }Processing time change event： the processing time of operation O3 belonging to job J4 is updated from 5 to 9.The current tasks： J1: { (O1, M2, PT1), (O2, M0, PT3), (O3, M1, PT6), (O4, M3, PT7), (O5, M5, PT3), (O6, M4, PT6) } J2: { (O1, M1, PT8), (O2, M2, PT5), (O3, M4, PT10), (O4, M5, PT10), (O5, M0, PT10), (O6, M3, PT4) } J3: { (O1, M2, PT10), (O2, M3, PT4), (O3, M5, PT8), (O4, M0, PT9), (O5, M1, PT1), (O6, M4, PT7) } J4: { (O1, M1, PT5), (O2, M0, PT5), (O3, M2, PT9), (O4, M3, PT3), (O5, M4, PT8), (O6, M5, PT9) } J5: { (O1, M2, PT9), (O2, M1, PT 3), (O3, M4, PT5), (O4, M5, PT4), (O5, M0, PT3), (O6, M3, PT1) } J6: { (O1, M1, PT3), (O2, M3, PT3), (O3, M5, PT9), (O4, M0, PT10), (O5, M4, PT4), (O6, M2, PT1) }Processing machine change event： the processing machine of operation O1 belonging to job J5 is updated from 2 to 0.The current tasks： J1: { (O1, M2, PT1), (O2, M0, PT3), (O3, M1, PT6), (O4, M3, PT7), (O5, M5, PT3), (O6, M4, PT6) } J2: { (O1, M1, PT8), (O2, M2, PT5), (O3, M4, PT10), (O4, M5, PT10), (O5, M0, PT10), (O6, M3, PT4) } J3: { (O1, M2, PT10), (O2, M3, PT4), (O3, M5, PT8), (O4, M0, PT9), (O5, M1, PT1), (O6, M4, PT7) } J4: { (O1, M1, PT5), (O2, M0, PT5), (O3, M2, PT9), (O4, M3, PT3), (O5, M4, PT8), (O6, M5, PT9) } J5: { (O1, M0, PT9), (O2, M1, PT3), (O3, M4, PT5), (O4, M5, PT4), (O5, M0, PT3), (O6, M3, PT1) } J6: { (O1, M1, PT3), (O2, M3, PT3), (O3, M5, PT9), (O4, M0, PT10), (O 5, M4, PT4), (O6, M2, PT1) }Processing machine change event： the processing machine of operation O6 belonging to job J6 is updated from 2 to 0.The current tasks： J1: { (O1, M2, PT1), (O2, M0, PT3), (O3, M1, PT6), (O4, M3, PT7), (O5, M5, PT3), (O6, M4, PT6) } J2: { (O1, M1, PT8), (O2, M2, PT5), (O3, M4, PT10), (O4, M5, PT10), (O5, M0, PT10), (O6, M3, PT4) } J3: { (O1, M2, PT10), (O2, M3, PT4), (O3, M5, PT8), (O4, M0, PT9), (O5, M1, PT1), (O6, M4, PT7) } J4: { (O1, M1, PT5), (O2, M0, PT5), (O3, M2, PT9), (O4, M3, PT3), (O5, M4, PT8), (O6, M5, PT9) } J5: { (O1, M0, PT9), (O2, M1, PT3), (O3, M4, PT5), (O4, M5, PT4), (O5, M0, PT3), (O6, M3, PT1) } J6: { (O1, M1, PT3), (O2, M3, PT3), (O3, M5, PT9), (O4, M0, PT10), (O 5, M4, PT4), (O6, M0, PT1) }The new tasks： J1: { (O1, M2, PT1), (O2, M0, PT3), (O3, M1, PT6), (O4, M3, PT7), (O5, M5, PT3), (O6, M4, PT6) } J2: { (O1, M1, PT8), (O2, M2, PT5), (O3, M4, PT10), (O4, M5, PT10), (O5, M0, PT10), (O6, M3, PT4) } J3: { (O1, M2, PT10), (O2, M3, PT4), (O3, M5, PT8), (O4, M0, PT9), (O5, M1, PT1), (O6, M4, PT7) } J4: { (O1, M1, PT5), (O2, M0, PT5), (O3, M2, PT9), (O4, M3, PT3), (O5, M4, PT8), (O6, M5, PT9) } J5: { (O1, M0, PT9), (O2, M1, PT3), (O3, M4, PT5), (O4, M5, PT4), (O5, M0, PT3), (O6, M3, PT1) } J6: { (O1, M1, PT3), (O2, M3, PT3), (O3, M5, PT9), (O4, M0, PT10),(O5, M4, PT4), (O6, M0, PT1) }

[unused17]The new tasks： J1: { (O1, M2, PT1), (O2, M0, PT3), (O3, M1, PT6), (O4, M3, PT7), (O5, M5, PT3), (O6, M4, PT6) }J2: { (O1, M1, PT8), (O2, M2, PT5), (O3, M4, PT10), (O4, M5, PT10), (O5, M0, PT10), (O6, M3, PT4) }J3: { (O1, M2, PT10), (O2, M3, PT4), (O3, M5, PT8), (O4, M0, PT9), (O5, M1, PT1), (O6, M4, PT7) }J4: { (O1, M1, PT5), (O2, M0, PT5), (O3, M2, PT9), (O4, M3, PT3), (O5, M4, PT8), (O6, M5, PT9) }J5: { (O1, M0, PT9), (O2, M1, PT3), (O3, M4, PT5), (O4, M5, PT4), (O5, M0, PT3), (O6, M3, PT1) }J6: { (O1, M1, PT3), (O2, M3, PT3), (O3, M5, PT9), (O4, M0, PT10), (O5, M4, PT4), (O6, M0, PT1) }The unavailable intervals of machines： M1: [(8, 11)]

Fig. 6. Results of slow-thinking

Furthermore, the model demonstrates promising generalization capability, suggesting that it can be extended to accommodate additional types of dynamic events without sacrificing solution interpretability or solver compatibility. Overall, these results underscore the practical applicability of the proposed fine-tuning approach for complex and dynamic scheduling environments.

## 5. Conclusion

This study introduces DScheLLM as one of the earliest demonstrations of large language models applied to dynamic job shop scheduling with adaptive reasoning capabilities. It addresses both known and previously unseen disruptions in dynamic production environments. By unifying natural-language reasoning with solver-level optimization, DScheLLM provides an extensible and interpretable alternative to traditional event-specific scheduling approaches, effectively bridging human-understandable decision-making and high-performance optimization. A central contribution of the proposed framework is to extend the dual-system (fast-slow) reasoning mechanism, which dynamically adapts to the scale and complexity of scheduling disruptions. The fast-thinking mode enables rapid local schedule adjustments, delivering feasible solutions in real time and achieving a 73.33% feasibility rate with 46.67% optimal solutions in fast-response scenarios. In contrast, the slow-thinking mode performs structured, stepwise reasoning over complex multi-event disturbances, generating standardized, solver-compatible representations that remain valid and interpretable across diverse combinations of dynamic events, thereby satisfying the precision requirements of industrial scheduling applications.

However, DScheLLM is not intended to replace exact optimization methods for large-scale combinatorial problems. Reasoning alone cannot reliably guarantee global optimality, and full-scale optimization continues to benefit from dedicated operations research solvers. Instead, the proposed framework enables automatic problem reasoning and modeling, substantially reducing reliance on manual intervention and expert knowledge, and thereby facilitating the rapid deployment and practical adoption of existing optimization models in real-world industrial settings. Future work will extend the framework to flexible job shops, flow shops, and multi-objective production settings, with the long-term objective of developing a general-purpose scheduling foundation model capable of reasoning and optimization across diverse industrial scenarios.

**Data availability statements**

The author confirms that all data generated during this study are included in this published article.